# Domain-Specific Shorthand for Generation Based on Context-Free Grammar

Andriy Kanyuka, akanyuka@tableau.com
Elias Mahfoud, emahfoud@tableau.com

## Abstract

The generation of structured data in formats such as JSON, YAML and XML is a critical task in Generative AI (GenAI) applications. These formats, while widely used, contain many redundant constructs that lead to inflated token usage. This inefficiency is particularly evident when employing large language models (LLMs) like GPT-4, where generating extensive structured data incurs increased latency and operational costs. We introduce a domain-specific shorthand (DSS) format, underpinned by a context-free grammar (CFG), and demonstrate its usage to reduce the number of tokens required for structured data generation. The method involves creating a shorthand notation that captures essential elements of the output schema with fewer tokens, ensuring it can be unambiguously converted to and from its verbose form. It employs a CFG to facilitate efficient shorthand generation by the LLM, and to create parsers to translate the shorthand back into standard structured formats. The application of our approach to data visualization with LLMs demonstrates a significant (3x to 5x) reduction in generated tokens, leading to significantly lower latency and cost. This paper outlines the development of the DSS and the accompanying CFG, and the implications of this approach for GenAI applications, presenting a scalable solution to the token inefficiency problem in structured data generation.

## Introduction

The generation of structured data is a fundamental task in Generative AI (GenAI) applications. These applications often rely on widely used formats such as JSON, YAML, and XML due to their versatility and compatibility across different systems. However, these general-purpose formats are characterized by a high degree of redundancy, leading to an inflated number of tokens required for data generation. This inefficiency becomes particularly pronounced when using large language models (LLMs) like GPT-4, where the generation of extensive structured data results in increased latency and operational costs.

To address this issue, our research introduces a domain-specific shorthand (DSS) format, used in conjunction with a context-free grammar (CFG), to significantly reduce the number of tokens required for structured data generation. This approach involves the creation of a shorthand notation that encapsulates the essential elements of the output schema with fewer tokens, while ensuring that it can be unambiguously converted to and from its verbose form.

We employ a CFG to facilitate the efficient generation of shorthand notation by the LLM, and to create parsers that can translate the shorthand back into standard structured formats. The application of our approach to the field of data visualization has demonstrated a significant (3x to 5x) reduction in token counts, leading to a proportional decrease in latency and cost.

This paper presents a detailed account of the development of the DSS and CFG, and the implications of this approach for GenAI applications. By addressing the token inefficiency problem in structured data generation, our research offers a scalable solution that can significantly enhance the performance and cost-effectiveness of GenAI applications.

In the following sections, we describe the specifics of our method, discuss the unique features of our approach, and present a practical application of our method for generating visualization specifications from natural language queries using an LLM. We also provide a review of related works in the field, highlighting the novelty and potential impact of our research.

## Related Work

The problem of generating structured data using large language models (LLMs) has been widely studied in the field of AI research. Several works have explored the use of context-free grammars (CFGs) and domain-specific languages (DSLs) to guide the generation process and ensure the output adheres to a specific format.

Allen-Zhu [1] studied how LLMs, such as GPT, learn CFGs. The study found that pre-trained transformers can generate sentences with near-perfect accuracy and impressive diversity for challenging CFGs. The hidden states of the transformers were found to implicitly encode the CFG structure, and the transformers could form "boundary to boundary" attentions that mimic dynamic programming.

Wang et al. [2] proposed grammar prompting, a method to enable LLMs to use external knowledge and domain-specific constraints, expressed through a grammar in Backus–Naur Form (BNF), during in-context learning. The method augments each demonstration example with a specialized grammar that is minimally sufficient for generating the particular output example. The LLM first predicts a BNF grammar given a test input, and then generates the output according to the rules of the grammar.

Willard and Louf [3] reformulated the problem of neural text generation in terms of transitions between the states of a finite-state machine. This approach allows the construction of an index over a language model's vocabulary, enabling the enforcement of domain-specific knowledge and constraints, and guaranteeing the structure of the generated text.

Ugare et al. [4] presented SynCode, a framework for efficient and general syntactical decoding with LLMs. SynCode leverages the CFG of a formal language, utilizing an offline-constructed efficient lookup table based on the discrete finite automaton (DFA) of the language grammar terminals. The framework was shown to eliminate all syntax errors and significantly outperform state-of-the-art baselines in generating JSON, Python, and Go outputs.

Beurer-Kellner et al. [5] introduced DOMINO, a novel decoding algorithm that can enforce constraints in a fully subword-aligned fashion, while leveraging pre-computation and speculative decoding to achieve virtually no overhead and in some cases even almost 2x speedup over unconstrained decoding.

Geng et al. [6] proposed a method for grammar-constrained decoding (GCD) that can serve as a unified framework for structured NLP tasks in general. They introduced input-dependent grammars, which allow the grammar to depend on the input and thus enable the generation of different output structures for different inputs.

Our work builds upon these studies by introducing a domain-specific shorthand (DSS) format guided by a CFG to reduce the token count required for structured data generation. This approach is unique in its focus on reducing the number of tokens generated, which directly impacts the latency and cost of GenAI applications. Our method has demonstrated significant improvements in efficiency, reducing the number of generated tokens by 78% in the provided example.

## Problem Statement

The generation of structured data is a fundamental task in Generative AI (GenAI) applications. These applications often utilize general-purpose formats such as JSON, YAML, and XML. However, these formats are characterized by a high degree of redundancy, leading to an inflated number of tokens required to generate them. The problem, therefore, lies in the inefficiency of these general-purpose formats in generating structured data, which leads to high latency and increased costs, especially when using larger models such as GPT-4. The challenge is to develop a method that can significantly reduce the number of tokens required for structured data generation, thereby reducing latency and cost.

## Proposed Solution

To mitigate the identified challenges, the proposed solution introduces a Domain-Specific Shorthand (DSS) in conjunction with a Context-Free Grammar (CFG) to enhance the efficiency of the generation process. By minimizing verbosity, DSS decreases the number of tokens required, leading to a reduction in generation time and associated costs. The shorthand notation is designed to encapsulate the critical elements of structured data succinctly. In many cases, elements defining the structure of the target format can be dropped or their number or lengths can be reduced significantly, leading to decrease of the overall document length, reducing the syntactic overhead and simplifying the generation process for large language models (LLMs).

Context-Free Grammar (CFG) is integral to the solution, providing a clear and efficient description of the shorthand notation to LLMs. CFG defines a set of production rules that specify how elements of the shorthand notation can be combined to produce valid structured outputs. These rules are minimal yet comprehensive, ensuring that the LLM can generate shorthand notation that is syntactically correct and semantically consistent with the domain-specific requirements. Additionally, CFG supports the development of parsers for translating shorthand notation back into full specification formats, maintaining interoperability and flexibility.

The development of a DSS begins with analyzing the general-purpose representation of structured data to identify opportunities for simplification and abstraction into a more concise notation. Following this, a CFG is developed to describe the shorthand notation in a manner understandable to an LLM, guiding it to generate shorthand that is both compact and adherent to the structure and semantics defined. Parsers and mappers are then created to facilitate the conversion between the DSS and the full specification formats, ensuring that the shorthand notation can be seamlessly integrated with existing systems and applications that utilize standard structured data formats.

Overall, the design and implementation of the DSS, underpinned by CFG, offer a streamlined and cost-effective method for generating structured data with LLMs. This approach addresses the challenges of latency and cost by reducing the token count, simplifying the generation process, and ensuring easy conversion between shorthand and full specification formats.

## Case Study

Our methodology was applied to the generation of data visualization specifications from user requests in natural language using large language models. These operations typically necessitate the LLM to output structured data, such as JSON, which serves as API payloads for accessing the underlying data sources and rendering the actual visualizations.

To achieve this, we first established a domain-specific schema tailored for data visualization (Appendix A). This schema captures essential visualization components such as fields, filters, and sorting criteria, which are common across various visualization types (e.g., bar charts, line graphs). Based on this schema, we designed a shorthand notation (Appendix B) that drastically reduced the verbosity inherent in JSON. For instance, the shorthand notation condensed complex JSON structures into concise, easily parsable lines of code, guided by a context-free grammar (CFG) specifically developed for this purpose (Appendix C).

The shorthand notation and CFG were designed with two primary objectives:
1. Token efficiency: the notation should significantly reduce the number of tokens required to represent visualization specifications.
2. Unambiguous mapping: it should allow for a clear and straightforward conversion back to the full JSON representation, ensuring no loss of information or ambiguity in the visualization requirements.

The shorthand notation not only achieved a significant reduction in token count but also maintained a high consistency with the original visualization specifications. Through the use of the CFG, the LLM was guided to generate shorthand notation that could be unambiguously parsed and converted back to the full JSON format, ensuring that the visualizations generated met the user's requirements accurately.

## Results and Discussion

Implementing a domain-specific shorthand (DSS) guided by context-free grammar (CFG) demonstrated a notable decrease in the token count necessary for representing structured data. In the conducted case study focused on visualization generation, the shorthand notation facilitated a 3x to 5x reduction in token counts when juxtaposed with the conventional JSON format. This reduction directly correlates with a comparable decrease in computational costs and latency, leading to enhanced response times for users and reduced operational expenses for services leveraging LLMs for dynamic visualization tasks.

The efficacy of the proposed methodology is underscored by its capacity to preserve the integrity of structured data while minimizing the resources required for its generation. The employment of CFG in the development of the shorthand notation ensures the data's definition remains clear and unambiguous, thereby simplifying the parsing process and enabling seamless conversion back to standard data formats. This advantage extends beyond mere cost and latency reductions, encompassing ease of implementation and system flexibility.

The broad applicability of this approach spans various domains necessitating the generation of structured data for API payloads, including but not limited to virtual assistants and agents, automated reporting and analytics, natural language interfaces for data processing, automation of unstructured data processing etc. The integration of DSS and CFG into structured data generation processes could significantly transform data management within GenAI applications, making it viable to deploy advanced models in scenarios previously hindered by cost or latency barriers. Moreover, this strategy may catalyze the creation of new domain-specific languages (DSLs) tailored to diverse applications, thereby improving the efficiency and accessibility of LLMs in structured data generation tasks.

## Conclusion

The research presented in this paper introduces a novel approach to structured data generation in Generative AI applications, addressing the inefficiency problem associated with the use of verbose, general-purpose formats such as JSON, YAML, and XML. By introducing a domain-specific shorthand (DSS) format, underpinned by a context-free grammar (CFG), we have demonstrated a significant reduction in the number of tokens required for structured data generation. This reduction directly impacts the latency and operational costs associated with the use of large language models like GPT-4, offering a scalable solution to the token inefficiency problem.

Our approach involves creating a shorthand notation that captures the essential elements of the output schema with fewer tokens, ensuring it can be unambiguously converted to and from its verbose form. The CFG facilitates efficient shorthand generation by the LLM and creates parsers to translate the shorthand back into standard structured formats. The application of our method to data visualization with LLMs demonstrated a 3x to 5x reduction in generated tokens, leading to significantly lower latency and cost.

The broad applicability of this approach spans various domains necessitating the generation of structured data for API payloads, including but not limited to virtual assistants and agents, automated reporting and analytics, natural language interfaces for data processing, and automation of unstructured data processing. The integration of DSS and CFG into structured data generation processes could significantly transform data management within GenAI applications, making it viable to deploy advanced models in scenarios previously hindered by cost or latency barriers. Moreover, this strategy may catalyze the creation of new domain-specific languages (DSLs) tailored to diverse applications, thereby improving the efficiency and accessibility of LLMs in structured data generation tasks.

In conclusion, our research presents a promising direction for future work in the field of GenAI applications, offering a practical and scalable solution to the token inefficiency problem in structured data generation.

## Appendix A: Example of a Full Spec

An example of an abstract visualization specification used in the case study as an API payload to render the viz. Length: 299 tokens

```
{
  "fields": [
    {
      "fieldName": "Order Date",
      "aggregation": "month",
      "encoding": "x",
      "role": "dimension",
      "type": "continuous",
      "dataType": "date"
    },
    {
      "fieldName": "Sales",
      "aggregation": "sum",
      "role": "measure",
      "type": "continuous",
      "dataType": "number"
    }
  ],
```

```
      "filters": [
        {
          "filterType": "categorical",
          "fieldName": "Product Name",
          "values": [
            "Product A",
            "Product B"
          ]
        },
        {
          "filterType": "categorical",
          "fieldName": "Region",
          "values": ["South", "West"]
        },
        {
          "filterType": "relative-date",
          "fieldName": "Order Date",
          "duration": 2,
          "units": "years"
        },
        {
          "filterType": "numeric-range",
          "fieldName": "Sales",
          "aggregation": "sum",
          "end": 10000,
          "start": 1000
        }
      ],
      "sort": [
        {
          "fieldName": "Region",
          "sortByField": "Sales",
          "aggregation": "sum",
          "direction": "desc",
          "limit": 5,
        }
      ]
    }
```

## Appendix B: Shorthand Example

The shorthand corresponding to the full spec in Appendix A, 66 tokens.

```
fields:
cd "Order Date" month x
cm "Sales" sum

filters:
cat "ProductName" values "Product A" "Product B"
rd "Order Date" 2 years
nr "Sales" sum start 1000 end 10000

sort:
"Sales" sum desc 5 "Region"
```

## Appendix C: Example of Context-free Grammar (CFG)

CFG used in the case study. It was included into the prompt to generate shorthand based on user queries in natural language. It was also used to create parsers to translate the shorthand to the full spec.

```
#CFG for high-level representation of dataset visualization in BNF
notation
#Lines starting with # are comments
#ε means empty string
<VizSpec> ::= <Fields><Filters><Sorting><ChartType>

#list of fields from the underlying dataset to be used in the
visualization
#may not include duplicates i.e fields with the same name
# must consist of the minimal subset of the dataset fields that satisfy
the user query
<Fields> ::= fields:\n <Field>...
<Field> ::= <FieldType> <FieldName> <Aggregation> <Encoding> \n

#FieldType should be continuous measure cm, continuous dimension cd or
discrete dimension dd
#insert or update it only if it's mentioned in user request, otherwise
leave as is
<FieldType> ::= cm | cd | dd | ε

#how the field values should be mapped to visual properties such as
color, size, shape and position
<Encoding> ::= color | size | shape | x | y | text | ε
```

```
<Aggregation> ::= count | countDistinct | sum | average | max | min |
median | year | quarter | month | week | day | hour | minute | second | ε

#filters to apply to the data used in the visualization
#non-empty only if user utterances mention filtering
<Filters> ::= filters:\n <Filter>... | ε

Filter ::=
<CategoricalFilter>|<RelativeDateFilter>|<DateRangeFilter>|<NumericalRang
eFilter>

#Categorical filter e.g. "show only banking and healthcare"
<CategoricalFilter> ::= cat <FieldName> <Exclude> values <FieldValues> \n
#Exclude: "show Segment except banking and healthcare"
Exclude ::= ex | ε

#relative date/time filter e.g. last year, previous two month
<RelativeDateFilter> ::= rd <FieldName><Units><Duration> \n
<Units> ::= days | weeks | months | quarters | years
<Duration> ::= number

#date range filter e.g. "between April and July" or "since last year"
<DateRangeFilter> ::= dr <FieldName> <StartDate> <EndDate> \n
#lower and upper bounds of the range
#missing start or end means "at most" or "at least" respectively
#ISO-8601-date-sring is a date in ISO 8601 format
<StartDate> ::= start <ISO-8601-date-sring>
<EndDate> ::= end <ISO-8601-date-sring>

#numeric range filter e.g. "values between 11 and 17"
# the field can be aggregated e.g. "field1" sum 10 20
<NumericalRangeFilter> ::= nr <FieldName> <Aggregation> <StartNumber>
<EndNumber> \n
#lower and upper bounds of the range
#missing start or end means "at most" or "at least" respectively
<StartNumber> ::= start <number>
<EndNumber> ::= end <number>

#how data in the visualization to be sorted
#should be present only if user utterances mention sorting
<Sorting> ::= sort:\n <Sort>... | ε

#the way data in the visualization to be sorted
```

```
#a field can be sorted individually, without affecting other fields in
the visualization
#"FieldName": field to sort, optional, should be used only if user
utterance mentions it
#SortByField: field to sort by
#Aggregation: aggregation to apply to sortByField e.g. sort Region by sum
of sales
#Direction: sort direction, ascending or descending
#Limit: how many items to include
<Sort>  ::= <SortByField><Aggregation><Direction><Limit><FieldName> \n
<Direction> ::= asc | desc | ε
<Limit> ::= <number> | ε

#type of the chart to be used in the visualization
#should be iuncluded only if user utterances mention chart type
<ChartType> ::=  chart:\n ChartTypeValue | ε
<ChartTypeValue> ::= text | heatmap | bar | stackedbar | line | area |
gantt | scatterplot | histogram | symbolmap | filledmap | treemap | pie

#pick field names and values from the supplied dataset extract
#put them in double quotes e.g. "Sales", "Furniture"
<FieldName>   ::= "quoted-field-name"
<SortByField> ::= "quoted-field-name"
<FieldValues> ::= "space-separated-quoted-string-values"
```